\documentclass{article}
\usepackage[preprint]{log_2023}			% for preprint version

\usepackage{booktabs}						% professional-quality tables
\usepackage{multirow}						% tabular cells spanning multiple rows
\usepackage{amsfonts}						% blackboard math symbols
\usepackage{graphicx}						% figures
\usepackage{duckuments}						% sample images
\usepackage[numbers,compress,sort]{natbib}	% for numerical citations
\usepackage{amsmath}
\usepackage{xcolor}
\definecolor{h}{rgb}{0.0, 0.56, 0.0}
\title[LCPool]{Graph Pooling by Local Cluster Selection}

\author[Y. Chen et al.]{%
Yizhu Chen\\ 
\email{cyzhaa@gmail.com}
}

\begin{document}

\maketitle

\begin{abstract}
Graph pooling is a family of operations which take graphs as input and produce shrinked graphs as output.
Modern graph pooling methods are trainable and, in general inserted in Graph Neural Networks (GNNs) architectures 
as graph shrinking operators along the (deep) processing pipeline.
This work proposes a novel procedure for pooling graphs, along with a node-centred graph pooling operator.

\end{abstract}

\section{Introduction}
Graph neural networks (GNNs) have shown their effectiveness in processing graph-structure data 
across lots of domains including social network, recommander system, drug discovery 
and time series\cite{gnnsurvey}\cite{DBLP:Recommendation}\cite{DBLP:moleculenet}\cite{DBLP:timeseriesimputation}. 
Graph pooling, as a way of hierarchically reducing graph representations, 
is crucial for capturing graph stuctures of different scales and coarsening large-scale graphs\cite{liu2023graphpoolinggraphneural}\cite{Grattarola_2024} 
with the purpose of applying neural graph processing to real-world problems\cite{diffpool}\cite{DBLP:rethinkpooling}. 
First graph pooling approaches are non-trainable, which use deterministic algorithms to pool graphs\cite{ndp}\cite{nmf}\cite{graclus}.
More advanced approaches are trainable and inserted in GNNs architectures as graph shrinking operators along the (deep) processing pipeline.\cite{DBLP:topk}\cite{sagpool}\cite{diffpool}\cite{mincutpool}\cite{asap}. 
Though many approaches are proposed which learn to pool graphs under different assumptions, 
they are similar in the way generating pooled graphs.\cite{Grattarola_2024}\cite{wang2024comprehensivegraphpoolingbenchmark}. 
This work proposes a new procedure for generating pooled graphs, which leverages the diversity
of modern graph pooling approaches. Along with the new approach, a node-centred graph pooling for average situations 
is proposed.

\section{Background}

A graph $G$ is a data structure described by a set of nodes $V$ and a set of edges 
$E$ that connect nodes. Node feature is denoted by $X \in \mathbb{R}^{\vert V \vert \times d}$, 
where $\vert V \vert$ is the number of nodes and $d$ the dimension of features. 
The edges is represented by adjacency matrix $A \in \mathcal{R^+} ^{\vert V \vert \times \vert V \vert}$, 
where entry $a_{i,j}$ is non-zero if there is an edge topologically associated with vertices $v_{i}$ and $v_{j}$. 

\begin{center}
	$ G = (V, E) = (X, A)$
\end{center} 

\textbf{Graph Convolution}. 
Graph convolution is the generalization of convolution operator to graph and updates the node features by convolution on graph\cite{gnnsurvey}, 
which can be based on spectral and spatial. 
Spatial graph convolution is formalized by message-passing(MP) scheme\cite{mpnn}, whose computation is

\begin{center}
 $x_i' = \gamma(h_i, x_i);$
 \quad
 $h_i = \oplus(\{m_{ij}| j\in \mathcal{N}(i) \});$
 \quad
 $m_{ij} = \phi(x_i, x_j, e_{ij})$
\end{center}

where $\gamma$ and $\phi$ are differentiable functions, $\oplus$ is a permutation-invariant and 
differentiable aggregation function, $x_i'$ is the updated node feature, $h_i$ and $m_{ij}$ are hidden features 
and $e_{ij}$ is the potential edge weight of edge connecting node $i$ and $j$.

\textbf{Graph Pooling}. A graph pooling operator $Pool(\cdot)$ generates a coarsened graph $G'$ starting form $G$ as 
\begin{center}
	$ G' = Pool(G) = (V', E') = (X', A')$
\end{center}
$V'$ and $E'$ represent the set of new nodes and edges.
$X'$ and $A'$ are the new node features and adjacency matrix. 
If $\vert V' \vert = 1$, then the pooling is named \textit{global pooling}, otherwise \textit{hierarchical pooling}.
In this paper, we only focus on \textit{hierarchical pooling}.

\textbf{SRC Pooling}. \citet{Grattarola_2024} propose a 
formal characterization of graph pooling based on three main operations, called \textit{selection}, \textit{reduction}, and \textit{connection}.
The selection function(SEL) computes $S=\{S_1, \ldots, S_K\}$, with each set $S_j={\{s_i^j|i \in [1,N]\} }$ representing the 
contributions of each original node to pooled node $j$. 
The reduction function (RED) calculates the new node features by $X'=RED(S,G)$ and 
the connection function (CON) calculates the new adjacency matrix by $A'=CON(S,G)$.

Graph poolings differentiate themselves based on the design of SEL, RED and CON functions. 
Though there are various SEL functions in the literature, 
the ways existing graph pooling methods generate pooled graphs (RED and CON functions) 
follow the same strategies. 
The most existing ways fall into two families, which are \textit{node selection} and \textit{dense assignment}.

\textbf{Node Selection}. Node selection is used by most sparse graph poolings\cite{Grattarola_2024}\cite{DBLP:topk}\cite{sagpool}, which generate the pooled graph by selecting some nodes 
and keeping edges between these nodes.
The selection is usually performed by ranking nodes with customed score function $g(\cdot)$ and select nodes with top $k$ scores
\begin{center}
	$h = g(X, A)$
	\\
	$\hat{i} = topk(h)$
	\\
	$X' = (X\odot h)(\hat{i},:)$
	\\
	$A' = A(\hat{i}, \hat{i})$
\end{center}
Where $topk(\cdot)$ is the function that returns index $\hat{i}$ of top $k$ inputs, 
$(\hat{i}, :)$ selects the $\hat{i}$ rows of a matrix and $(\hat{i}, \hat{i})$ selects 
the $\hat{i}$ rows and $\hat{i}$ columns of a matrix.

\textbf{Dense Assignment}.
Dense assignment is used by most dense pooling approaches\cite{Grattarola_2024}\cite{diffpool}\cite{mincutpool} 
and learns a dense matrix $S$ with a fixed second dimension for input graph $G$ by an assignment function $f(\cdot)$ 
that generates pooled graph $G'$ by matrix multiplication
\begin{center}
	$S = f(X, A)$
	\\
	$X' = S^TX$
	\\
	$A' = S^TAS$
\end{center}
The main difference of existing graph pooling approaches is 
in the score function $g(\cdot)$ or the assignment function $f(\cdot)$. 

\section{Local Cluster Pooling}
We propose a new way of generating pooled graphs with following advantages: 
1) Our method is adaptive \cite{Grattarola_2024} and computes pooled graphs that have a size proportional to that of the input;
2) Our method can utilize sparse matrix multiplication to reduce computational cost;
3) Our method strengthen the connectivity of pooled graph by constructing new edges.
Then we show how it simplifies under common conditions. Finally, we propose a graph pooling learns
to pool graph from both node feature and their differences among local neighbours.

\subsection{Local Assignment Selection}

\textit{Node selection} and \textit{dense assignment} processes introduce some inherent properties for pooling approaches adpoting them. 
The \textit{node selection} only filter edges without constructing new ones, 
resulting in inappropriate connections in some cases.
The \textit{dense assignment} derives from inherently dense matrix $S$ with a fixed second dimension, 
therefore related matrix multiplications 
are computationally costly and the pooled graphs are of fixed size, 
which causes several small graphs to be upscaled and very big graphs to be excessively shrunk\cite{Grattarola_2024}. 
Besides, the dense adjacency matrix $A'$ of pooled graph requires extra efforts to be made when 
cooperate the graph poolings with sparse graph convolution layers. 

We propose \textit{local assignment selection} as follows.
Given a graph $G=(X, A)$, $S$ is the assignment matrix learnt by an assignment fuction $f(\cdot)$, $g(\cdot)$ is a score function.
If $f(\cdot)$ satisfies prerequisite \textbf{1} and the pooled graph $G'=(X', A')$ is generated in the following way, 
then we call this way \textit{local assignment selection}.

\textbf{Prerequisite 1}.
$S \in \mathcal{R}^{|V|\times |V|}$ is an assignment matrix produced by $f(\cdot)$. $A^*=I_N+A$ is the adjacency matrix with self-loop. 
$s_{ij}$ representing the contribution of node $i$ to node $j$ 
is non-zero \textit{only when} $a_{ij}$ is not zero.

We call such $S$ the \textit{local assignment matrix}. 
The prerequisite \textbf{1} requests that each node \textit{can only} contribute to its neighbours (and itself).
The properties of \textit{local assignment matrix} brings three benefits.
1). The assignment matrix is not inherently dense 
and could utilize sparse matrix operations to reduce computational cost;
2). The assignment is localized so that disconnected nodes are not pooled together;
3). The second dimension of $S$ is not fixed (depending on the dimension of $A$), so the size of pooled graph is flexible;

\textit{Local assignment selection} generates pooled graphs as

\begin{center}
	$S=f(X,A)$
	\\
	$X^* = S^TX$
	\\
	$h = g(X^*,A)$
	\\
	$\hat{i} = topk(h)$
	\\
	$X' = (X^*\odot h)(\hat{i},:)$
	\\
	$S' = S(:,\hat{i})$
	\\
	$A' = {S'}^TAS'$
\end{center}

\begin{figure}[h]
	\includegraphics[width=\textwidth]{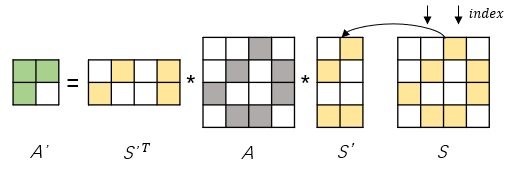}
	\caption{The figure shows how the adjacency matrix of pooled graph is calculated.}
	\label{fig:edge_rebuild}
\end{figure}

Where $(:,i)$ selects the columns $\hat{i}$ of matrix and $(\hat{i},:)$ selects the rows. 
The node features are updated by local assignment matrix first, 
then $g(\cdot)$ calculates the scores $h$ of updated features. 
Nodes are selected by these scores and edges are constructed by sparse matrix multiplication. 
$h$ is multiplied to node features to train $g(\cdot)$ jointly.
The figure \ref{fig:edge_rebuild} shows a visualization of how new adjacency matrix $A'$ is calculated.

In this way, the connectivity of pooled graphs are strengthened compared to 
ones produced by \textit{node selection}. Precisely, we have the proposition \textbf{2}. 
The node features are also more representative since they are calculated by the aggregating 
contributors. 

\textbf{Proposition 2}.
\textit{For coarsened graphs produced by local assignment selection, 
edges are constructed between pooled nodes if they have connected contributors, 
and the edges of original graph are preserved if all the diagonal entries $s_{i,i}$ of $S$ are not zero.}

The way ASAPool\cite{asap} generates pooled graph can be cast in our framework; here $f(\cdot)$ is an attention output and $g(\cdot)$ a customed MP layer. Considering 
the process \textit{local assignment section} and \textit{node selection} have in common, the later 
could be viewed as a class of instances of the former.

\textbf{Proposition 3}.
\textit{Node selection could be viewed as local assignment selection with $f(\cdot)$ is a costant function: $f(\cdot) = I_N$}.

\subsection{Local Cluster Selection}

A common class of $f(\cdot)$ requests each node \textit{must} contribute to its neighbours (and itself). 
We say such kind of $f(\cdot)$ satisfies prerequisite \textbf{4}.

\textbf{Prerequisite 4}.
$S \in \mathcal{R}^{|V|\times |V|}$ is an assignment matrix produced by $f(\cdot)$. $A^*=I_N+A$ is the adjacency matrix with self-loop. 
$s_{ij}$ representing the contribution of node $i$ to node $j$ 
is non-zero \textit{when and only when} $a_{ij}$ is not zero.

For such a $f(\cdot)$, \textit{local assignment selection} can produce coarsened
graphs without calculating detailed assignment matrix $S$.
We call it \textit{local cluster selection}, in which a cluster function $v(\cdot)$ takes place of $f(\cdot)$.
Below lemma \textbf{5} gives the description of \textit{local cluster selection}, dismissing the assignment matrix $S$.

\textbf{Lemma 5}. 
\textit{For a function $v(\cdot)$ updating node features, if:}

\begin{enumerate}
	\item \textit{$v(\cdot)$ updates feature of node $i$ with exactly itself and its one-hop neighbours.}
	\item \textit{The original and coarsened graphs have unweighted edges.}
\end{enumerate}

\textit{Then local assignment selection simplifies as}

\begin{center}
	$X^* = v(X,A)$
	\\
	$h = g(X^*,A)$
	\\
	$\hat{i} = topk(h)$
	\\
	$X' = (X^* \odot h)(\hat{i},:)$
	\\
	$A' = ones(A + A^TA + A^2 + A^TA^2)(\hat{i}, \hat{i})$
\end{center} 
\textit{Where $ones(\cdot)$ is the function that replaces all non-zero entries of matrix with constant 1.}

Notably, if the graph is undirected, $A$ is symmetrical and
proposition \textbf{6} holds.

\textbf{Proposition 6}. \textit{For undirected graphs, 
$A' = ones(A + A^2 + A^3)(\hat{i}, \hat{i})$}.

\subsection{Local Cluster Pooling}
We propose local cluster pooling (LCPool) that pools graphs by local cluster selection. 
Since $v(\cdot)$ updates node features in the same way 
that graph convolutions perform, we could use any convolution operation satisfying the condition of lemma \textbf{5} as $v(\cdot)$.
Considering the convolution parts of a GNN model before pooling, we could just dismiss function $v(\cdot)$ in the pooling process.

\textbf{Proposition 7}. Function $v(\cdot)$ could be dismissed if the convolution layer 
before pooling satisfies the condition in lemma \textbf{5}.

We validate proposition \textbf{7} by experiments. 
For LCPool, we design a function $g(\cdot)$ which learns the scores from both node features and 
differences among local neighbours to select the clusters. 

\subsubsection{From graph laplacian to LCSMP}

Given a graph $G$ with adjacency matrix $A$, Its graph laplacian matrix is
$L = D-A$, 
where $D$ is the degree matrix of the graph. 
Graph laplacian could be viewed as a discrete laplacian operator on a graph. 
By multiplying graph laplacian and node features $X$, 	
$H^* = LX$. Expanding the matrix multiplication, the value of an entry $h^*_{i,j}$ of matrix $H^*$ is given by:

\begin{center}
	$h^*_{i,j} = x_{i,j}*d_i - \sum_{k \in \mathbf{N}(i)}x_{k,j}$
\end{center}  

Where $d_i$ is the degree of node $i$ and $\mathbf{N}(i)$ is the set formed by neighbours of node $i$. 
Collecting all the columns, we have the feature vector $\tilde{h^*_i}$ of node $i$.

\begin{center}
	$\tilde{h^*_i} = \tilde{x_i}*d_i - \sum_{k \in \mathbf{N}(i)}\tilde{x_k}$
\end{center}

Where $\tilde{x_i}$ is the feature vector of node $i$. 
Since the cardinality of $N(i)$ is $d_i$, we could rewrite the equation in following formula.

\begin{center}
	$\tilde{h^*_i} = \sum_{k \in \mathbf{N}(i)}(\tilde{x_i}- \tilde{x_k})$
\end{center}

So actually the multiplication of $L$ and $X$ performs an aggregation of node differences between local neighbours 
and produces a feature matrix $H^*$ which represents the differences. 
If we multiply feature vector $\tilde{h^*_i}$ of node $i$ by a learnable vector $\tilde{w}$, 
we get a function $g^*$ that learns score from the node differences among local neighbours of node $i$.

\begin{center}
	$g^*(\tilde{x_i})=(\sum_{k \in \mathbf{N}(i)}(\tilde{x_i}- \tilde{x_k}))\tilde{w}^T $
\end{center} 

\begin{figure}[h]
	\centering
	\includegraphics[width=0.8\textwidth]{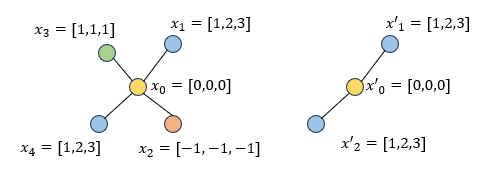}
	\caption{$g^*(\cdot)$ can not distinct $x_0$ and $x'_0$ in the two local clusters above, since $x_0-x_3$ and $x_0-x_2$
	cancels each other out.}
	\label{fig:cancel}
\end{figure}

The problem here is that the differences cancel each other out if we do not perform any operation before summation, 
which limites the distinction ability of function $g^*(\cdot)$, see figure \ref{fig:cancel}.

So we introduce another learnable weight matrix to transform the difference $(\tilde{x_i}- \tilde{x_k})$ 
before we sum them up.  
To simplify the notation, we use linear layer to denote the operation that transforms the features. 
The definition of linear layer is given as follows.

\textbf{Linear layer.} \textit{For an input feature matrix $X \in \mathcal{R}^{N\times F_1}$, a linear layer $L$ transforms $X$ in the way 
$L = XW^T+\tilde{b}$, where $W \in \mathcal{R}^{F_2 \times F_1}$ is the learnable weight matrix 
and $\tilde{b} \in \mathcal{R}^{1\times F_2}$ 
is the learnable bias. $N$ is the number of features, $F_1$ is the dimension of input features and $F_2$ is the dimension of output features.}

Then our score function $g$ produces the score $h_i$ of node feature $x_i$ in the following way.

\begin{center}
	$h_i = L_s (L_{f_d}(\sum_{k \in \mathbf{N}(i)}L_d(\tilde{x_i}- \tilde{x_k})) + L_x(\tilde{x_i}))$
\end{center}

Nonlinearity is added after each linear layer. We use \textit{softmax} as the final nonlinearity after $L_s$, and 
\textit{relu} as the internal nonlinearity for other linear layers.
The feature differences of node $i$ and its neighbours are first transformed into hidden features by linear layer $L_d$ 
, so that they do not just cancel each other out, 
then their aggregation are transform by $L_{f_d}$ into hidden feature that represents the difference among the local neighbours of node $i$.
This hidden feature is summed with the feature learnt from $x_i$ by $L_x$ to learn also from node feature itself. 
After that, a linear layer $L_s$ learns the score $h_i$ of node $i$ from summed feature. 

\begin{figure}[h]
	\centering
	\includegraphics[width=0.9\textwidth]{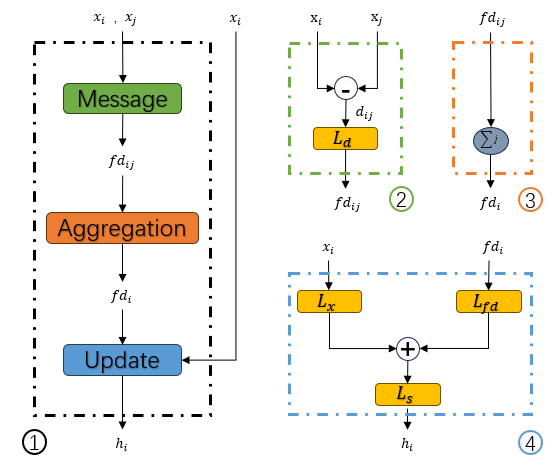}
	\caption{This figure shows the architecture of LCSMP. $L$ denotes the linear transformation, 
	the unfilled circle denotes mathematical operation and filled cirecle denote the sum aggregation through neighbours. The subfigure 1 shows the overall structure of LCSMP from a message passing perspective.
	The subfigure 2 shows the message part of LCSMP. We calculate the differences and transform them with a linear layer.
	The subfigure 3 shows the aggregation part. 
	The subfigure 4 shows how the importance score is learned from hidden features.}
	\label{fig:LcsMP}
\end{figure}

This function could be implemented as an MP layer which we call it local cluster score MP(LCSMP).
The figure \ref{fig:LcsMP} shows the architecture of LCSMP layer.

\subsubsection{Equation description of LCPool}

We assumpt that the input and coarsened graphs are undirected and have unweighted edges.
With score function $g(\cdot)=\text{LCSMP}(\cdot)$, LCPool could be described as

\begin{center}
	$h = \text{LCSMP}(X,A)$
	\\
	$\hat{i} = topk(h)$
	\\
	$X' = (X \odot h)(\hat{i},:)$
	\\
	$A' = ones(A+A^2+A^3)(\hat{i},\hat{i}) $
\end{center}

where LCSMP$(\cdot)$ is given by

\begin{center}
	$\text{LCSMP}(\tilde{x_i})= L_s (L_{f_d}(\sum_{k \in \mathbf{N}(i)}L_d(\tilde{x_i}- \tilde{x_k})) + L_x(\tilde{x_i}))$
\end{center}

\section{Experiment}
\subsection{Goals of Experiment}

The experiment is mainly designed having in mind two tasks.
\begin{enumerate}
	\item Our proposed LCPool is a competitive and effective graph pooling approach in common situations.
	\item Our proposition \textbf{7} of dismissing the cluster function $v(\cdot)$ of \textit{local cluster section} is valid.
\end{enumerate}

The most common task solved by GNNs is graph classification. We build two GNN backbones for graph classification.
The classification accuracy of backbones with different pooling approaches shows the effectiveness of pooling approaches.
Our proposed pooling approach is compared with several existing approaches to see whether its competitive and effective.
A variant \textbf{LCPool*} of our method is given by taking an extra GCN layer as $v(\cdot)$. 
The variant pooling operator is evaluated to see whether dismissing $v(\cdot)$ impedes 
the pooling effectiveness.

\subsection{GNN Backbones}
The two GNN backbones utilize coarsened graph representations in hierarchical and plain style, and
they consist of four kinds of components: Multi-layer perceptrons(MLPs), graph convolution blocks, graph pooling operators and readout layer. 
MLPs transform vector features before and after GNN parts. Graph convolution blocks perform convolution on graph representations.
Graph pooling operators shrink graph representations. Readout layer reduces graph representations into vector features. 
Following previous work\cite{sagpool}, The readout function is defined as 
\begin{center}
	$\text{readout}(X) = \text{concat}(\text{global\_mean}(X), \text{global\_max}(X))$
\end{center}
where global\_mean$(\cdot)$ takes the mean of $X$, global\_max$(\cdot)$ take the maximum of $X$ 
and concat$(\cdot)$ is the concatenation function.

The GNN backbones assemble the four components as figure \ref{fig:GNN} shows. The detailed settings of components are given in appendix.

\begin{figure}[h]
	\centering
	\includegraphics[width=0.8\textwidth]{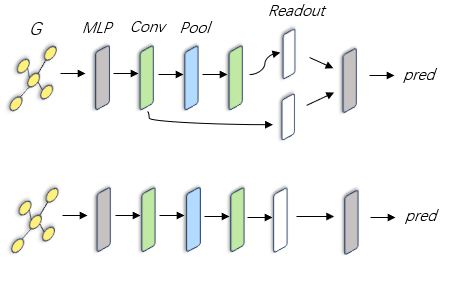}
	\caption{The figure shows the architecture of GNN backbones. The top one is in 
	hierarchical style and the bottom one in plain style. The grey shape denotes MLP, 
	green shape denotes graph convolution operator, blue shape denotes graph pooling operator and 
	unfilled shape denotes readout layer.}
	\label{fig:GNN}
\end{figure}

\subsection{Baseline Approaches} 

We chose several node-centred graph pooling approaches for comparison. 
Some of them pool graphs by \textit{node selection} (TopKPool\cite{DBLP:topk}, SAGPool\cite{sagpool}), 
while others pool graphs by \textit{dense assignment}(DiffPool\cite{diffpool}, MincutPool\cite{mincutpool}).
ASAPool\cite{asap} pools graph by \textit{local assignment selection}. 
NoPool denotes backbones without pooling operators.

\subsection{Graph Classification Datasets}

We chose the common classification datasets used by previous works. 
They are on different domains and available from TUDataset\cite{tudataset}, including 
PROTEINS, ENZYMES, Mutagenicity, DD, NCI1 and COX2. 
Parts of their properties of are summarized as table \ref{table:dataset}.

\begin{table}
	\caption{Summary of datasets.}
	\centering
	\begin{tabular}{llllllll}
	\hline

	& \textit{Graphs} & \textit{Classes} & \textit{Avg. Nodes} & \textit{Avg. Edges} & \textit{Node Labels} & \textit{Node Attr.}& \\ \hline 
	PROTEINS& 1113 &  2 &  39.06 & 72.82  & +  & +(1)  &\\
	ENZYMES&  600 & 6  & 32.63  &  62.14 & +  & +(18)  &\\
	Mutagenicity&  4337 &  2 &  30.32 & 30.77  & +  & -  &\\
	DD&  1178 & 2  & 284.32  & 715.66  & + &  - &\\
	NCI1&  4110 & 2  & 29.87  & 32.30  & +  &  - &\\
	COX2&  467 &   2&   41.22&   43.45&   +&   +(3)&\\

	\end{tabular}
	\label{table:dataset}
\end{table}

\subsection{Results}
For every model, we evaluate it 10 runs and the results are summarized as tables. 
We train the model by early-stoping with patience, 
and the datasets are divided into training, validation and testing datasets with a ratio of 8:1:1. 
The hyperparameters are consistent for every pooling approaches and they are tuned so that 
every model converge and there is no apparently overfitting (model stops trainning at first epoch). 
See detailed settings in appendix.

The tables \ref{table:classifier_h} and \ref{table:classifier_hc} show the evaluation results of hierarchical backbones,
where $\text{GNN}_h$ denotes the hierarchical backbone with GCN and $\text{GNN}_h^c$ denotes the one with GraphConv.
The tables \ref{table:classifier_p} and \ref{table:classifier_pc} show results of plain backbones, 
where $\text{GNN}_p$ denotes the plain backbone with GCN and $\text{GNN}_p^c$ denotes the one with GraphConv.
For each task, pooling approach achieving the highest accuracy is highlighted, excluding the variant lcpool*.
We observe that no single pooling approach outperforms the others systematically, 
and for different backbones, the approaches achieving highest accuracy for given tasks vary, 
which indicates that the chosing of best pooling approach depending on the combination of backbones and tasks.
However, the results show that our approach is \textbf{more competitive and expressive in general}. 
It achieves four highest accuracies for different backbones and tasks, which is the most.
And for the situation our approach is not the best, it still ranks among the top several. 
To clarify this claim, for every backbone, we rank the accuracy of each pooling approach on every tasks 
and produce an average ranking for that approach. The rankings are summarized as table \ref{table:ranking}.
We can see that our approach ranks highest for every backbone, 
which supports the claim our approach is competitive and expressive. 
We also observe that the variant lcpool* with $v(\cdot)= \text(GCN)(\cdot)$  does not outperform our method, 
which validates our \textbf{proposition \textbf{7} of dismissing the cluster function $v(\cdot)$}. 

\begin{table}[htbp]
	\caption{Accuracy of hierarchical backbone with GCNConv.}
	\footnotesize
	\centering
	\begin{tabular}{llllllll}
	\hline
	$\text{GNN}_h$ & PROTEINS & ENZYMES  & Muatagenicity & DD & NCI1 & COX2 & \\ \hline
	nopool &75.00\tiny{{$\pm${2.00}}}  & 70.17\tiny{{$\pm${5.19}}} &78.02\tiny{{$\pm${1.63}}} &            73.56\tiny{{$\pm${1.19}}}    &76.98\tiny{{$\pm${3.15}}}&82.77\tiny{{$\pm${2.60}}}& \\ 
	topkpool&74.73\tiny{{$\pm${2.00}}}  & 65.00\tiny{{$\pm${4.08}}}&  77.95\tiny{{$\pm${1.21}}}       &       75.51\tiny{{$\pm${1.58}}} &  77.64\tiny{{$\pm${2.32}}}&  82.77\tiny{{$\pm${4.89}}}&\\
	sagpool & 74.11\tiny{{$\pm${1.87}}} & 65.50\tiny{{$\pm${4.02}}} & 78.13\tiny{{$\pm${1.81}}}      &          \textcolor{h}{75.93\tiny{{$\pm${2.12}}}} & \textcolor{h}{79.78\tiny{{$\pm${2.03}}}} & 84.89\tiny{{$\pm${2.42}}} & \\ 
	asapool& 73.57\tiny{{$\pm${1.92}}}  & -                         & \textcolor{h}{80.09\tiny{{$\pm${1.42}}}}      &       75.00\tiny{{$\pm${1.98}}} &            79.00\tiny{{$\pm${2.52}}} & 84.47\tiny{{$\pm${3.57}}} & \\ 
	diffpool& 73.84\tiny{{$\pm${1.74}}} & 71.00\tiny{{$\pm${3.18}}} & 79.22\tiny{{$\pm${0.67}}} & 72.37\tiny{{$\pm${1.38}}} &    74.62\tiny{{$\pm${1.84}}} & 84.47\tiny{{$\pm${4.47}}} & \\ 
	mincutpool& 74.82\tiny{{$\pm${2.45}}} & \textcolor{h}{71.17}\tiny{{$\pm${4.48}}} & 79.12\tiny{{$\pm${1.01}}} & 73.14\tiny{{$\pm${2.89}}} &   76.16\tiny{{$\pm${2.22}}}  & 83.19\tiny{{$\pm${2.93}}}  &\\\hline 
	lcpool&  \textcolor{h}{75.71\tiny{{$\pm${1.25}}}}&66.67\tiny{{$\pm${3.16}}}  &79.52\tiny{{$\pm${1.31}}}  &  74.15\tiny{{$\pm${1.79}}}&       79.10\tiny{{$\pm${1.75}}} & \textcolor{h}{85.69\tiny{{$\pm${1.70}}}} & \\ 
	$\text{lcpool}^*$&  74.46\tiny{{$\pm${1.14}}}&64.17\tiny{{$\pm${3.10}}}  &79.40\tiny{{$\pm${0.86}}}  &  73.05\tiny{{$\pm${4.23}}}&           81.80\tiny{{$\pm${1.86}}} & 83.83\tiny{{$\pm${3.46}}} & \\ 
	\end{tabular}
	\label{table:classifier_h}
\end{table}

\begin{table}[htbp]
	\caption{Accuracy of hierarchical backbone with GraphConv.}
	\footnotesize
	\begin{tabular}{llllllll}
	\hline
	$\text{GNN}_h^c$ & PROTEINS & ENZYMES  & Muatagenicity & DD & NCI1 & COX2 &  \\ \hline
	nopool &74.02\tiny{{$\pm${1.67}}}& 73.50\tiny{{$\pm${3.37}}} &80.21\tiny{{$\pm${0.90}}}&\textcolor{h}{77.46\tiny{{$\pm${1.94}}}}&\textcolor{h}{81.63\tiny{{$\pm${1.79}}}}&\textcolor{h}{85.32\tiny{{$\pm${2.01}}}}&  \\ 
	topkpool&73.48\tiny{{$\pm${1.50}}} & 71.67\tiny{{$\pm${5.63}}}&  80.32\tiny{{$\pm${2.28}}}&  76.27\tiny{{$\pm${0.93}}}&  79.78\tiny{{$\pm${1.64}}}&  83.19\tiny{{$\pm${3.49}}}& \\
	sagpool & \textcolor{h}{75.27\tiny{{$\pm${0.90}}}} & 71.67\tiny{{$\pm${3.80}}} & 79.12\tiny{{$\pm${1.60}}} & 75.00\tiny{{$\pm${2.30}}} & 80.68\tiny{{$\pm${1.73}}} & 85.10\tiny{{$\pm${4.76}}} &  \\ 
	asapool& 73.48\tiny{{$\pm${2.77}}} & 77.00\tiny{{$\pm${3.93}}} & 79.86\tiny{{$\pm${1.45}}}  & 76.10\tiny{{$\pm${2.30}}} & 78.73\tiny{{$\pm${1.37}}} & 81.91\tiny{{$\pm${3.33}}} &  \\ 
	diffpool& 73.66\tiny{{$\pm${1.21}}} & \textcolor{h}{79.83\tiny{{$\pm${3.11}}}} & 79.40\tiny{{$\pm${0.96}}} & 76.53\tiny{{$\pm${1.78}}} & 76.98\tiny{{$\pm${1.98}}} & 81.70\tiny{{$\pm${2.17}}} &  \\ 
	mincutpool& 72.77\tiny{{$\pm${1.80}}} & 75.67\tiny{{$\pm${3.89}}} & 79.84\tiny{{$\pm${0.98}}} & 76.10\tiny{{$\pm${2.93}}} &77.20\tiny{{$\pm${1.57}}}  & 78.30\tiny{{$\pm${4.34}}}  &  \\\hline
	lcpool&  74.91\tiny{{$\pm${1.89}}}&74.50\tiny{{$\pm${2.36}}}  &\textcolor{h}{80.51\tiny{{$\pm${1.20}}}}  &  76.10\tiny{{$\pm${3.10}}}& 78.61\tiny{{$\pm${1.98}}} & 84.04\tiny{{$\pm${2.38}}} &  \\ 
	$\text{lcpool}^*$&  74.55\tiny{{$\pm${2.23}}}&74.83\tiny{{$\pm${2.92}}}  &81.61\tiny{{$\pm${1.32}}}  &  72.88\tiny{{$\pm${2.14}}}& 81.00\tiny{{$\pm${1.76}}} & 83.19\tiny{{$\pm${3.35}}} &  \\ 
	\end{tabular}
	\label{table:classifier_hc}
\end{table}

\begin{table}[htbp]
	\caption{Accuracy of plain backbone with GCNConv.}
	\footnotesize
	\centering
	\begin{tabular}{llllllll}
	\hline
	$\text{GNN}_p$ & PROTEINS & ENZYMES  & Muatagenicity & DD & NCI1 & COX2 &  \\ \hline
	nopool &75.27\tiny{{$\pm${3.32}}}& 64.17\tiny{{$\pm${2.61}}} &79.24\tiny{{$\pm${1.07}}}&72.88\tiny{{$\pm${3.19}}}&79.68\tiny{{$\pm${2.85}}}&82.55\tiny{{$\pm${4.54}}}&  \\ 
	topkpool&\textcolor{h}{75.54\tiny{{$\pm${2.00}}}} & 60.33\tiny{{$\pm${3.64}}}&  76.41\tiny{{$\pm${1.44}}}&  71.02\tiny{{$\pm${2.33}}}&  79.44\tiny{{$\pm${1.06}}}&  77.23\tiny{{$\pm${4.95}}}& \\
	sagpool & 73.93\tiny{{$\pm${1.99}}} & 63.17\tiny{{$\pm${3.98}}} & 79.31\tiny{{$\pm${1.20}}} & \textcolor{h}{77.54\tiny{{$\pm${2.61}}}} & 76.20\tiny{{$\pm${2.20}}} & 81.28\tiny{{$\pm${2.48}}} &  \\ 
	asapool& 74.29\tiny{{$\pm${2.07}}} & 62.67\tiny{{$\pm${4.48}}} & \textcolor{h}{80.16\tiny{{$\pm${1.20}}}}  & 72.54\tiny{{$\pm${4.39}}} & \textcolor{h}{79.76\tiny{{$\pm${1.95}}}} & 81.91\tiny{{$\pm${3.59}}} &  \\ 
	diffpool& 72.77\tiny{{$\pm${1.51}}} & \textcolor{h}{71.33\tiny{{$\pm${3.23}}}} & 79.33\tiny{{$\pm${1.18}}} & 72.71\tiny{{$\pm${1.64}}} & 75.06\tiny{{$\pm${1.53}}} & 82.98\tiny{{$\pm${2.69}}} &  \\ 
	mincutpool& 72.68\tiny{{$\pm${2.68}}} & 67.17\tiny{{$\pm${4.95}}} & 80.05\tiny{{$\pm${0.75}}} & 74.07\tiny{{$\pm${3.33}}} &75.16\tiny{{$\pm${1.68}}}  & \textcolor{h}{83.83\tiny{{$\pm${3.04}}}}  & \\\hline 
	lcpool&  72.68\tiny{{$\pm${3.39}}}&67.67\tiny{{$\pm${3.00}}}  &79.47\tiny{{$\pm${0.68}}}  &  74.32\tiny{{$\pm${2.28}}}& 79.49\tiny{{$\pm${1.37}}} & 82.98\tiny{{$\pm${1.65}}} &  \\ 
	$\text{lcpool}^*$&  72.86\tiny{{$\pm${2.16}}}&66.17\tiny{{$\pm${3.25}}}  &79.24\tiny{{$\pm${0.83}}}  &  75.08\tiny{{$\pm${3.08}}}& 81.80\tiny{{$\pm${2.11}}} & 80.00\tiny{{$\pm${3.95}}} &  \\ 
	\end{tabular}
	\label{table:classifier_p}
\end{table}

\begin{table}[htbp]
	\caption{Accuracy of plain backbone with GraphConv.}
	\footnotesize
	\centering
	\begin{tabular}{llllllll}
	\hline
	$\text{GNN}_p^c$ & PROTEINS & ENZYMES  & Muatagenicity & DD & NCI1 & COX2 &  \\ \hline
	nopool &73.57\tiny{{$\pm${1.96}}}& 71.83\tiny{{$\pm${4.56}}} &80.44\tiny{{$\pm${0.74}}}&75.08\tiny{{$\pm${3.01}}}&\textcolor{h}{81.63\tiny{{$\pm${1.93}}}}&80.64\tiny{{$\pm${4.51}}}&  \\ 
	topkpool&73.30\tiny{{$\pm${2.48}}} & 70.17\tiny{{$\pm${5.94}}}&  78.92\tiny{{$\pm${1.26}}}&  72.80\tiny{{$\pm${1.99}}}&  80.58\tiny{{$\pm${1.86}}}&  \textcolor{h}{83.19\tiny{{$\pm${2.93}}}}& \\
	sagpool & 73.21\tiny{{$\pm${1.87}}} & 74.33\tiny{{$\pm${4.36}}} & 79.91\tiny{{$\pm${1.33}}} & 74.66\tiny{{$\pm${3.27}}} & 80.63\tiny{{$\pm${1.99}}} & 82.77\tiny{{$\pm${3.74}}} &  \\ 
	asapool& 73.21\tiny{{$\pm${2.68}}} & 69.83\tiny{{$\pm${5.18}}} & \textcolor{h}{80.67\tiny{{$\pm${1.44}}}}  & 71.10\tiny{{$\pm${5.16}}} & 79.03\tiny{{$\pm${1.57}}} & 81.49\tiny{{$\pm${4.15}}} &  \\ 
	diffpool& 72.95\tiny{{$\pm${2.53}}} & \textcolor{h}{76.17\tiny{{$\pm${3.88}}}} & 79.54\tiny{{$\pm${0.99}}} & \textcolor{h}{76.69\tiny{{$\pm${3.01}}}} & 77.79\tiny{{$\pm${2.40}}} & 82.55\tiny{{$\pm${4.13}}} &  \\ 
	mincutpool& 71.96\tiny{{$\pm${2.83}}} & 74.17\tiny{{$\pm${4.03}}} & 80.00\tiny{{$\pm${1.25}}} & 76.36\tiny{{$\pm${1.99}}} &78.56\tiny{{$\pm${1.53}}}  & 80.64\tiny{{$\pm${3.22}}}  & \\\hline 
	lcpool&  \textcolor{h}{74.02\tiny{{$\pm${1.85}}}}&71.83\tiny{{$\pm${6.69}}}  &80.48\tiny{{$\pm${0.94}}}  &  73.39\tiny{{$\pm${3.42}}}& 80.61\tiny{{$\pm${2.17}}} & 82.77\tiny{{$\pm${4.30}}} &  \\ 
	$\text{lcpool}^*$&  73.30\tiny{{$\pm${1.57}}}&73.50\tiny{{$\pm${3.53}}}  &80.83\tiny{{$\pm${1.29}}}  &  70.51\tiny{{$\pm${3.12}}}& 81.65\tiny{{$\pm${1.30}}} & 81.91\tiny{{$\pm${3.95}}} &  \\ 
	\end{tabular}
	\label{table:classifier_pc}
\end{table}

\begin{table}[htbp]
	\caption{Ranking of pooling approaches with different backbones.}
	\footnotesize
	\centering
	\begin{tabular}{lllllllll}
	\hline
	\textit{Ranking} & nopool & topkpool  & sagpool & asapool & diffpool & mincutpool & lcpool&\\ \hline
	$\text{GNN}_h$ &   4.67&  4.67 & 3.33  &  3.40   &  4.83   &  4.17   &  \textcolor{h}{2.33}   &\\
	$\text{GNN}_h^c$ &   \textcolor{h}{2.33}&  4.00 & 4.17  &  4.17  & 4.33    & 5.5   & 3.5   &\\
	$\text{GNN}_p$ &   3.67&  5.5 & 4.33  & 3.67    &  4.17   & 3.5    & \textcolor{h}{3.17}    &\\
	$\text{GNN}_p^c$ &   3.33&  4.5 & 3.17  &  5.00   &  4.17   & 4.67    & \textcolor{h}{3.17}    &\\
	\end{tabular}
	\label{table:ranking}
\end{table}

\section{Conclusion}

In this work, we propose a framework of generating pooled graphs for 
advanced and trainable graph pooling approaches. 
Our method provides following important benefits for graph pooling approaches adpoting it:
1) Adaptive;
2) Utilizing sparse matrix multiplication;
3) Strengthening the connectivity of pooled graphs.
We proof that one of the existing frameworks (\textit{node selection}) can be viewed as a subset of our framework.
Besides, we proof the framework can simplify under some common conditions. 
Along with the simplified framework, 
we propose a novel graph pooling operator that further simplifies the pooling process
and pools graphs by node attributes and their differences among local neighbours.
The experiment shows our pooling operator is competitive and expressive, 
and its proposition of simplification is valid.

% For natbib users:
\bibliographystyle{unsrtnat}
\bibliography{reference}
% For bibLaTeX users:
% \printbibliography

\appendix
\section{Proofs}
\textbf{Lemma 8}. \textit{Let $S$ denote the local assignment matrix and $A$ the adjacency matrix. Operator $(\hat{i},:)$ selects $\hat{i}$ rows of matrix, 
$(:,\hat{i})$ selects $\hat{i}$ columns of matrix and $(\hat{i}, \hat{i})$ selects the rows $\hat{i}$ and columns $\hat{i}$ of matrix, we have the following equation vaild.
\begin{center}
	${S(:,\hat{i})}^TA(S(:,\hat{i})) = (S^TAS)(\hat{i},\hat{i})$
\end{center}
}

\subsection{Proof of Proposition 2}

According to lemma \textbf{8}, we have $A' = (S^TAS)(\hat{i},\hat{i})$. 
For node $p$ and $q$, if they share connected contributors, then there exsits 
node $n$ and $m$ such that (\textit{s.t.}) node $n$ contributes to node $p$, node $m$ 
contributes to node $q$ and node $m$ and node $n$ are connected. 
Then we have $s^t_{p,n}$, $a_{n,m}$ and $s_{m,q}$ are non-zero, where $s^t$ denotes the entry of $S^T$.
Let $H=S^TAS$, expanding the matrix multiplication, we have $h_{p,q}$ is non-zero. 
So there is an edge between node $p$ and node $q$ in pooled graph. 

If the diagonal entries $s_{i,i}$ of S are all non-zero, then each node must contribute to itself. 
Considering two node $p$ and $q$ which are connected in original graph, they have the connected contributors $p$ and $q$,
so that there is an edge constructed between them in pooled graph. In the other word, the edge connecting 
node $p$ and $q$ are preserved if we don't consider the edge weights. 

\subsection{Proof of Proposition 3}
Substitute $f(\cdot) = I_N$ into the equations of local assignment selection, we have

\begin{center}
	$X^* = X$
	\\
	$h = g(X^*, A) = g(X, A)$
	\\
	$\hat{i} = topk(h)$
	\\
	$X' = (X^*\odot h)(\hat{i}, ;) = (X\odot h)(\hat{i}, ;)$
	\\
	$A' = {I_N(\hat{i},:)}^TA(I_N(:,\hat{i})) = (I_N^TAI_N)(\hat{i}, \hat{i}) = A(\hat{i}, \hat{i}) $
\end{center}
which exactly describes the node selection process. 

\subsection{Proof of Lemma 5}

Since we don't care about the edge weights, we could just replaces $S$ with $I_N+A$, which doesn't change 
the positions of non-zero entries. 
Since we don't acer about edge weights, we have $A'=ones(S'^TAS')$.
According to the lemma \textbf{8}, we have
\begin{center}
	$A'=ones(S'^TAS') = ones({S(:,\hat{i})}^TA(S(:,\hat{i}))) = ones((S^TAS)(\hat{i},\hat{i}))$
\end{center}
Replacing $S$ with $I_N+A$, we have 
\begin{center}
	$A' = ones((I_N+A)^TA(I_N+A)(\hat{i},\hat{i}))$
\end{center}
Expanding the equations, we have
\begin{center}
	$A' = ones((I_N^TAI_N+A^TAI_N+I_N^TAA+A^TAA)(\hat{i},\hat{i})) = ones((A+A^TA+A^2+A^TA^2)(\hat{i},\hat{i}))$
\end{center}
We could take the rows and columns selection operator out, since it does not change the value of entries.
Then we have
\begin{center}
	$A' = ones(A+A^TA+A^2+A^TA^2)(\hat{i},\hat{i})$
\end{center}
Then the value of entries of $S$ is only used to calculate $X^*$, so we could integrate 
the equations 
\begin{center}
	$S = f(X, A)$
	\\
	$X^* = S^TX$
\end{center}
into one function $v(\cdot)$ \textit{s.t.} $X^* = v(X,A)$, where $v(\cdot)$ satisfies the conditions.

\subsection{Proof of Proposition 6}
If graphs are undirected, we have $A$ which is symmetrical, then $A^T=A$. Substitute the equation,
we have 
\begin{center}
	$A' = ones(A+A^TA+A^2+A^TA^2)(\hat{i},\hat{i}) = ones(A+2*A^2+A^3)(\hat{i},\hat{i})$
\end{center}
since $ones(\cdot)$ replaces all non-zero entries with one, 
\begin{center}
	$A' = ones(A+2*A^2+A^3)(\hat{i},\hat{i}) =ones(A+A^2+A^3)(\hat{i},\hat{i}))$
\end{center}

\section{Experiment settings}
We training the models using early-stoping with patience. The patience is set to $50$ and the 
maximum epoches is set to $500$. The batch size is $32$ and the learning rate is $0.0005$. 
The dimension of hidden features is $128$, 
while the pre-MLP has the linear layers of size $[128]$ and the post-MLP has the 
linear layers of size $[256, 128]$. We use ReLU as the nonlinearity in the backbones. 
Activations of pooling operators are set as their default ones.

\end{document}